\documentclass[fleqn,10pt,twocolumn]{SICE23}

\usepackage{amsmath}

\title{Speed Optimization Algorithm based on Deterministic Markov Decision Process for Automated Highway Merge}

\author{Takeru Goto${}^{1\dagger}$, Kosuke Toda${}^{1}$ and Takayasu Kumano${}^{1}$}
\speaker{Takeru Goto}

\affils{${}^{1}$Innovative Research Excellence, Honda R\&D Co., Ltd., Tokyo, Japan\\
(Tel: +81-80-9456-0182; E-mail: takeru\_goto@jp.honda)\\
}

\abstract{%
This study presents a robust optimization algorithm for automated highway merge.
The merging scenario is one of the challenging scenes in automated driving,
because it requires adjusting ego vehicle's speed to match other vehicles before reaching the end point.
Then, we model the speed planning problem as a deterministic Markov decision process.
The proposed scheme is able to compute each state value of the process and reliably derive the optimal sequence of actions.
In our approach, we adopt jerk as the action of the process to prevent a sudden change of acceleration.
However, since this expands the state space, we also consider ways to achieve a real-time operation.
We compared our scheme with a simple algorithm with the Intelligent Driver Model.
We not only evaluated the scheme in a simulation environment but also conduct a real world testing.
 }
\keywords{%
Markov Decision Process, Dynamic Programming, Automated Driving, Highway Merge
}

\begin{document}

\maketitle


\section{Introduction}
In recent years, a lot of researches are conducted to realize automated driving.
In 2021, the first vehicle equipped with a Level 3 automated driving system for low-speed ranges on highways was launched by Honda. However, it does not support merging at ramps which is one of the challenging scenarios in automated driving. The scenario needs the speed planner to adjust the ego vehicle's speed to other vehicles by the end of the merging lane. To tackle this, a number of schemes are considered, such as machine learning based control\cite{learning1,learning2}, model predictive control (MPC)\cite{mpc1,mpc2} and cooperative control with vehicle-vehicle (v2v) communication\cite{connect1,connect2}. Machine learning based approaches can generate optimal actions in scenarios where it is difficult to build a model in advance, but ensuring performance is yet challenging. MPC predicts a future state and minimizing a value function to obtain the optimal control value by solving a quadratic programming problem. It also does not guarantee that the solution is optimal due to the time constraint. Cooperative controls realize the absolute safe and comfortable behavior using accurate information from each other. However, all vechiles must have v2v communication systems, though it is difficult in the near future.

Therefore, we model the speed planning problem as deterministic Markov decision process (MDP)\cite{sutton}. MDP is a basis theory of reinforcement learning when the model of state transition is unknown.
However, if the model is known, we can guarantee obtaining the optimal solution by calculating state values.
On the other hand, it tends to require a numerous of computation because of the state's size.
Then, we design the appropriate state space and introduce some calculation techniques to compute in real-time.
Moreover, we evaluate our scheme in not only a simulation environment, but also a real test course.

\section{Speed Optimization based on deterministic MDP}
\subsection{Modeling}
Markov decision process is a stochastic model for dynamic systems with decision making. It is defined by the state space $S$, the action space $A(s)$, the state transition probability $P(s'|s,a)$, and the reward function $r(s,a,s')$ where $s,s'\in{S}$ and $a\in{A(s)}$. In deterministic MDP, when action $a$ is selected in state $s$, the transition from s to s' is guaranteed. Therefore, we consider the state transition function $f:S\times{A}\to{S}$ instead of $P(s'|s,a)$.

We adopt jerk as the action. Therefore, action space $A$ is as 
\begin{eqnarray}
\begin{array}{rcl}
    A&=&{\lbrace}j_{-n_j},\ldots,j_0,j_1,\ldots,j_{n_j}\rbrace \\
    &=&{\lbrace}-n_j\varDelta{j},\ldots,0,\varDelta{j},\ldots,n_j\varDelta{j}\rbrace, \label{eq:action_space}
\end{array}
\end{eqnarray}
where $\varDelta{j}$ is a quantization step of jerk and $n_j$ is a value determined by the maximum jerk.
The state is composed of time, trajectory position, velocity and acceleration. The set of time $T$ is described as
\begin{eqnarray}
\begin{array}{rcl}
    T&=&{\lbrace}t_0,t_1,\ldots,t_{n_v}\rbrace \\
    &=&{\lbrace}0,\varDelta{t},\ldots,n_t\varDelta{t}\rbrace, \label{eq:time_space}
\end{array}
\end{eqnarray}
where $\varDelta{t}$ is a discrete time step and $n_t$ is a value determined by the maximum planning time.
The quantization steps of them except time depend on $\varDelta{j}$ and $\varDelta{t}$ because of the equation of motion.
Each set of acceleration, velocity and trajectory is respectively as
\begin{eqnarray}
\begin{array}{rcl}
    G&=&{\lbrace}g_{-n_g},\ldots,g_0,\ldots,g_{n_g}\rbrace \\
    &=&{\lbrace}-n_g\varDelta{j}\varDelta{t},\ldots,0,\ldots,n_g\varDelta{j}\varDelta{t}\rbrace, \label{eq:accl_space}
\end{array}
\end{eqnarray}
\begin{eqnarray}
\begin{array}{rcl}
    V&=&{\lbrace}v_0,v_1,\ldots,v_{n_v}\rbrace \\
    &=&{\lbrace}0,\varDelta{j}\varDelta{t}^2/2,\ldots,n_v\varDelta{j}\varDelta{t}^2/2\rbrace, \label{eq:velocity_space}
\end{array}
\end{eqnarray}
\begin{eqnarray}
\begin{array}{rcl}
    L&=&{\lbrace}l_0,l_1,\ldots,l_{n_l}\rbrace \\
    &=&{\lbrace}0,\varDelta{j}\varDelta{t}^3/6,\ldots,n_v\varDelta{j}\varDelta{t}^3/6\rbrace, \label{eq:position_space}
\end{array}
\end{eqnarray}
where $n_g$, $n_v$ and $n_l$ are the values determined by the maximum acceleration, velocity and trajectory position respectively.
Then, the state space is expressed as $S=T\times{G}\times{V}\times{L}$. Eqs. (\ref{eq:time_space}) to (\ref{eq:position_space}) show the state space is inversely propostional to the senventh power of $\varDelta{t}$ and the forth power of $\varDelta{j}$. Therefore, we need to decide these values with consideration for computational cost. 

\begin{figure}[t]
\centering
\includegraphics[width=8cm]{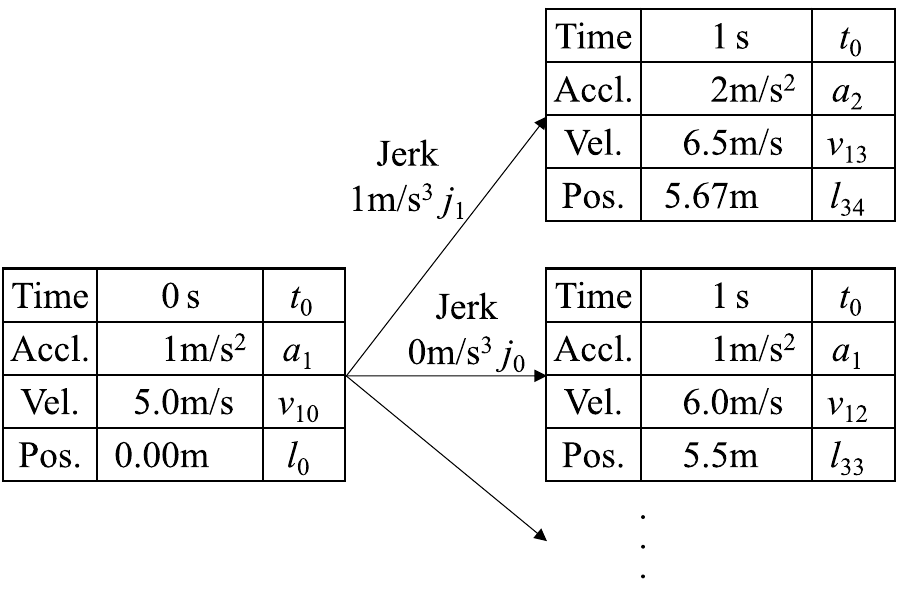}
\caption{Example of the state transition.}
\label{fig:state_transition}
\end{figure}

The rule of the state transition from a state $s=[t_{i_t}, g_{i_j}, v_{i_v}, l_{i_l}]$ when the selected action is $j_{i_j}$ is derived from the equation of motion.
We illustrate the example of the state transition when $\varDelta{t}=1\mathrm{s}$ and $\varDelta{j}=1\mathrm{m/s^3}$ in Fig. \ref{fig:state_transition}. In general, the state transition function $f$ is expressed as
\begin{multline}
f(s, j_{i_j})\\=[t_{i_t+1},g_{i_g+i_j},v_{i_v+2i_g+i_j},l_{i_l+3i_v+3i_g+i_j}].
\end{multline}

\subsection{State Value and Reward Function}
The action value which is the sum of future rewards when $a$ is selected as an action in the state s is as
\begin{equation}
Q(s,a)=r(s,a,f(s,a))+\gamma{V(f(s,a))}, \label{eq:action_value}
\end{equation}
where $V$ is the value function of $s$ and $\gamma<1$ is a decay rate to prevent the state value from divergence.
We also adopt the deterministic greedy policy similar to the state transition.
Therefore, the policy function and the state value function is expressed by Eq. \ref{eq:action_value} as
\begin{equation}
\pi(s)=\underset{a}{\operatorname{argmax}}\lbrace{Q(s,a)}\rbrace,\\ \label{eq:policy}
\end{equation}
\begin{equation}
V(s)=\underset{a}{\operatorname{max}}\lbrace{V(f(s,a))}\rbrace \quad(s\notin{S_T}), \label{eq:state_value} 
\end{equation}
where $S_T$ is a set of termination state.

\begin{figure}[t]
\centering
\includegraphics[width=8cm]{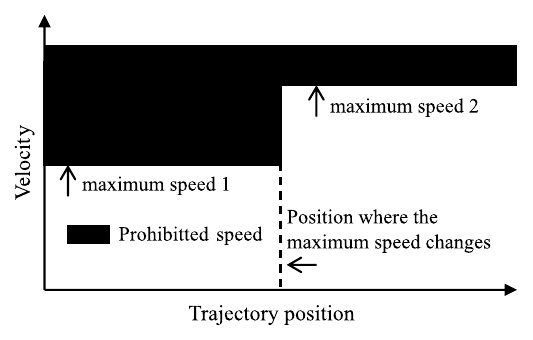}
\caption{The state which is terminated by the limit speed.}
\label{fig:vel_and_pos}
\end{figure}

\begin{figure}[t]
\centering
\includegraphics[width=8cm]{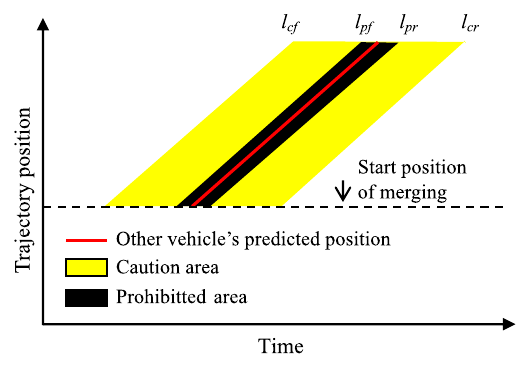}
\caption{The state which is terminated and the state whose reward is attenuated by other vehicles.}
\label{fig:pos_and_time}
\end{figure}

Next, we design the reward function and the value of termination state considering the ego vehicle's state, the action and the distance from other vehicles.
There are two kinds of termination state. The first is the state of the end of time, i.e. the time $t$ in the state is $n_t\varDelta{t}$. The second is a state that is prohibited due to safety reasons or traffic rules.
We put these states as $S_{T_e}$ and $S_{T_p}$ respectively. Therefore, $S_T=S_{T_e}\cup{S_{T_p}}$.
Fig. \ref{fig:pos_and_time} shows the safety and danger area because of another vehicle on the target lane.
$l_{cf\_i}$, $l_{pf\_i}$, $l_{pr\_i}$ and $l_{cr\_i}$ are the function of $t$ based on the other vehicle's predicted position.
$l_{pf\_i}$ and $l_{pr\_i}$ express the absolute safety merge from another vehicle $i$.
$l_{cf\_i}$ and $l_{cr\_i}$ are the ideal distance between ego vehicle and the vehicle $i$.
We use a simple linear prediction currently, though other efficient prediction algorithms can be used.
Fig. \ref{fig:vel_and_pos} illustrates the limit speed $v_{max}(l)$ according to the position.
$S_{T_p}$ is defined simply in this paper considering the distance from other vehicles and the limit speed as 
\begin{equation}
\begin{split}
S_{T_p}=&\bigcup_{i<n_{o}}{\left\{t\in{T},l\in{L}|l_{pr\_i}(t)\le l \le l_{pf\_i}(t)\right\}}\\
&\qquad \cup\left\{l\in{L},v\in{V}|v_{max}(l)<v\right\},\label{eq:tp} 
\end{split}
\end{equation}
where $n_o$ is a number of other vehicles.
We define the basic functions which form the reward and the value function of the termination state following as
\begin{equation}
r_a(a)=\exp\left\{-\left(\frac{j}{n_j\varDelta{j}}\right)^2\right\},\\ \label{eq:r_a}
\end{equation}
\begin{equation}
\begin{split}
r_s(s)=&\exp{\left\{-\left(\frac{a}{n_a\varDelta{a}}\right)^2\right\}}\times\\
&\qquad\exp{\left\{-\left(\frac{v-v_{max}(l)}{v_{max}(l)}\right)^2\right\}}.\label{eq:r_s}
\end{split}
\end{equation}
These equations means that acceleration and jerk should be lower, and the velocity should be near the limit speed at that position.
The current functions are expressed by exponential function. However, if the value is greater than 0 and 1 or less, we can adopt other shape functions. 
We introduce the attenuate function for the ego vehicle to break out of the caution area in Fig. \ref{fig:pos_and_time}.
\begin{equation}
att(s,i)=
\begin{cases}
\frac{l-l_{cr\_i}}{l_{pr\_i}-l_{cr\_i}} & (l_{cr\_i}\le l< l_{pr\_i})\\
\frac{l-l_{pf\_i}}{l_{cf\_i}-l_{pf\_i}} & (l_{pf\_i}< l\le l_{cf\_i})\\
1 & (l<l_{cr\_i},l_{cf\_i}<l).
\end{cases}\label{eq:att} 
\end{equation}
From Eqs. (\ref{eq:r_a}), (\ref{eq:r_s}) and (\ref{eq:att}), the reward function is designed as
\begin{equation}
r(s,a,s')=
\begin{cases}
\begin{split}
&\underset{i}{\operatorname{min}}\left\{att(s',i)\right\}\times\\
&\qquad r_a(a)\times{r_s(s')}.
\end{split}
& (V(s')\ne0)\\
0 & (V(s')=0),
\end{cases}\label{eq:reward}
\end{equation}
The value of the termination state is as
\begin{equation}
V(s)=
\begin{cases}
\frac{1}{1-\gamma}r_s(s) & (s\in{S_{T_e} \: \text{and} \: s\notin{S_{T_p}}})\\
0 & (s\in{S_{T_p}}).
\end{cases}\label{eq:terminate_value} 
\end{equation}
The coefficient $1/(1-\gamma)$ has a roll to equalize the scale of the state value among different times\cite{reward}.
The value of the state inevitably reaching the termination state $s$ of $S_{T_p}$ must be zero from Eqs. (\ref{eq:reward}) and (\ref{eq:terminate_value}).
The value of other states is greater than zero similar to it.
It guarantees that the scheme can absolutely avoid the collision and breaking rules if there is a solution.

\subsection{Calculation techniques}
$V$ is a recurrence formula by Eq. \ref{eq:state_value}.
Furthermore, states with $t_{i}$ must transition to states with $t_{i+1}$.
Thus, the value of all states can be calculated in reverse from the end of time without dynamic programming.
This method does not require complex branching such as recursive processing, and is easy to parallelize.
However, it also calculates for states that cannot be reached from the initial state. Therefore, the calculation is optimized by limiting the calculation target to the position and velocity reachable from the initial state's velocity and acceleration.

If the $\varDelta{t}$ is less than 1, it is obvious that the size of the position space would become huge from Eq. (\ref{eq:position_space}). Thus, we thin out the elements in the position space. It is not recommended that the acceleration and the speed are thinned out, since they affects the result of the position integrally.

\section{Experiments}
\subsection{Evaluation in simulation}
\begin{figure}[t]
\centering
\includegraphics[width=8cm]{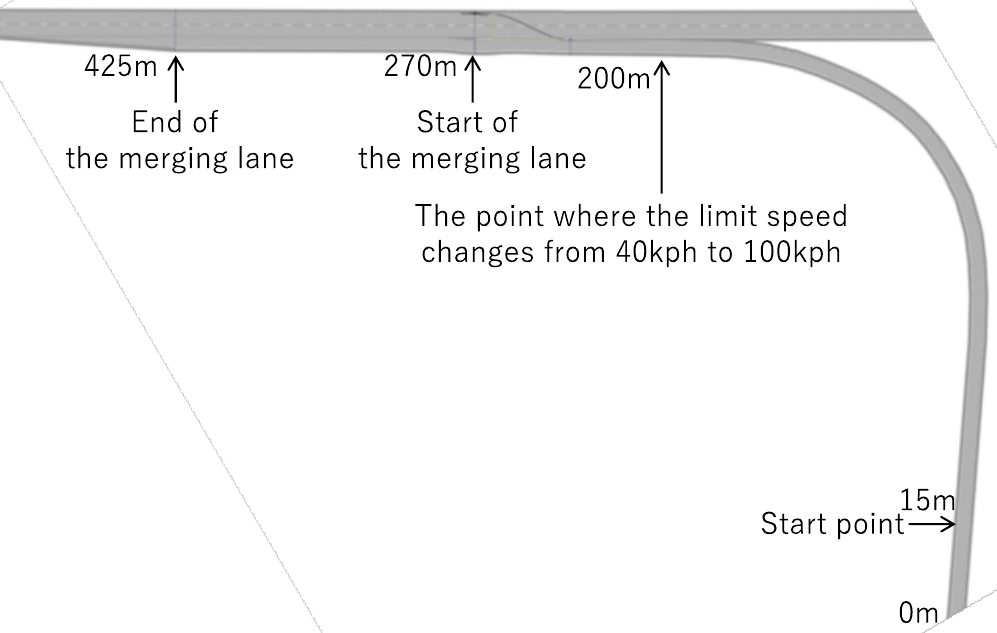}
\caption{The course where the merging test is conducted.}
\label{fig:test_course}
\end{figure}

\begin{figure}[t]
\centering
\includegraphics[width=8cm]{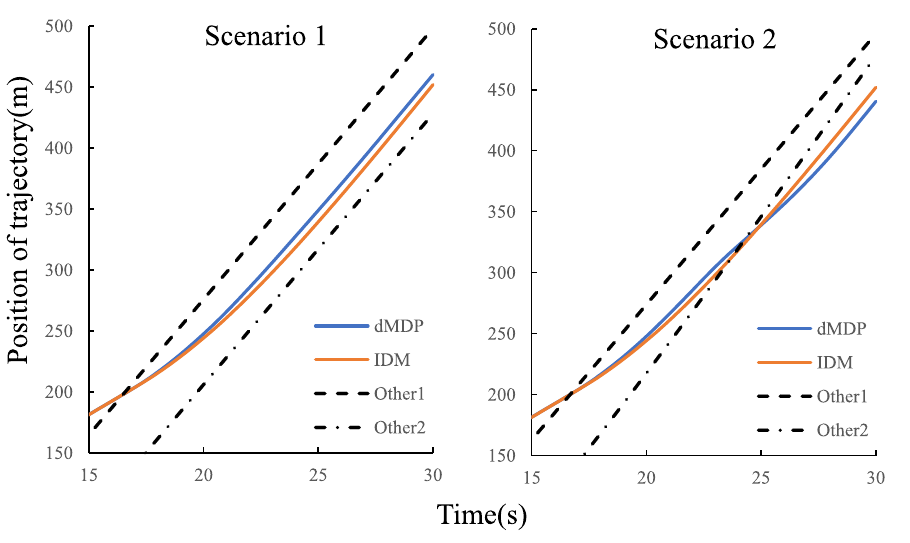}
\caption{The result of position of trajectory.}
\label{fig:position}
\end{figure}

\begin{figure}[t]
\centering
\includegraphics[width=8cm]{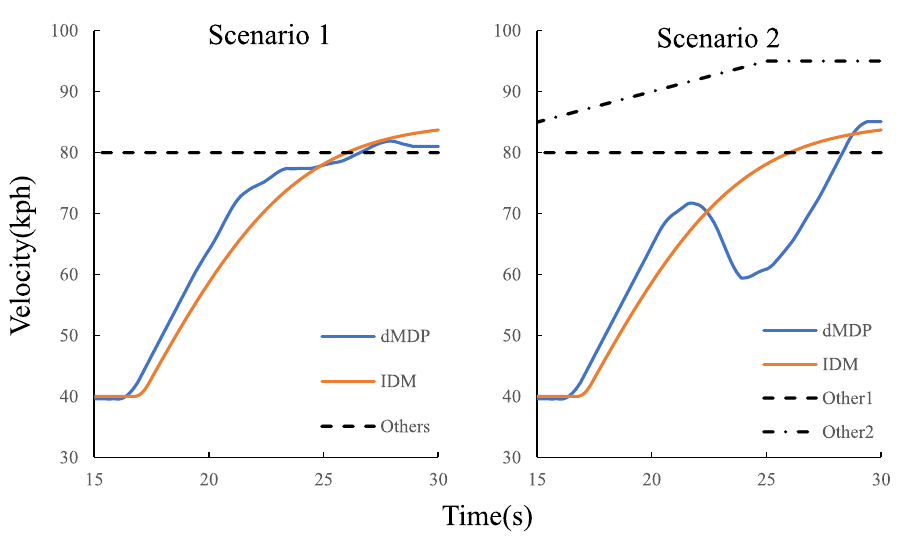}
\caption{The result of velocity.}
\label{fig:velocity}
\end{figure}

Fig. \ref{fig:test_course} illustrates the shape of the test course. The model of the course is the high speed oval course at Japan Safe Driving Center Central Training Academy for Safe Driving in Ibaraki, Japan.
There are two scenarios. Two other vehicles keep constant speed at 80 kph in the scenario 1. The first vehicle keeps 80 kph, but the second vehicle accelerates at 95 kph to hinder the ego vehicle in the scenario 2.
We introduce a simple planner with Intelligent Driver Model\cite{idm} which is for simulating the human's following motion to compare with the proposed scheme.

Fig. \ref{fig:position} and Fig. \ref{fig:velocity} show the result of position and velocity respectively.
In scenario 1, the IDM based planner is closer to the second vehicle than the proposed algorithm, because it doesn't take the rear vehicle in account. The proposed scheme consider all other vehicles simultaneously.
Therefore, the MDP based planner can take the ego vehicle into the balanced position.
In scneario 2, while our algorithm has the temporal deceleration after accelerating, the relative position from the second vehicle is appropriate when merging.
In fact, the IDM planner generates the smooth speed. However, the most important thing in ramps is to ensure the appropriate relative distance with other vehicles. We have not conducted enough evaluation yet. These results indicate the proposed scheme is superior to simple following schemes.

\subsection{Implementation in real test vehicle}
\begin{table}[tb]
\caption{Calculation time of speed planning.}
\label{tb:calc_time}
\begin{center}
\begin{tabular}{|c|c|c|c|}\hline
&Min.&Max.&Avg. \\\hline
Calc. time(ms)& 3.1 & 23.3 & 8.1 \\\hline
\end{tabular}
\end{center}
\end{table}
The calculation is conducted in a PC equipped with Intel Core i9 10980HK.
We parallelize the calculation with 8 threads.
Tbl. \ref{tb:calc_time} is the result of calculation time.
The calculation time is increased when there is no other vehicles and the speed limit is high, because the situation has the least number of the termination states.
In any case. the calculation is finished with the planning step time which is $0.1\mathrm{s}$.

\begin{figure}[t]
\centering
\includegraphics[width=8cm]{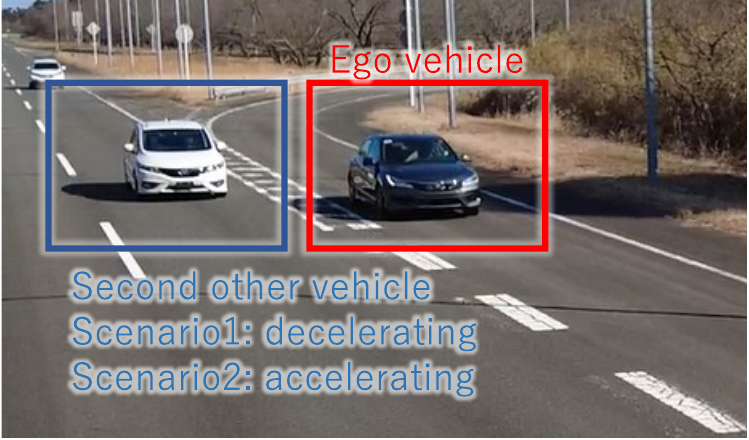}
\caption{The real world test. The scenario is that the second vehicle is accelerating.}
\label{fig:real_test_scene}
\end{figure}

\begin{figure}[t]
\centering
\includegraphics[width=8cm]{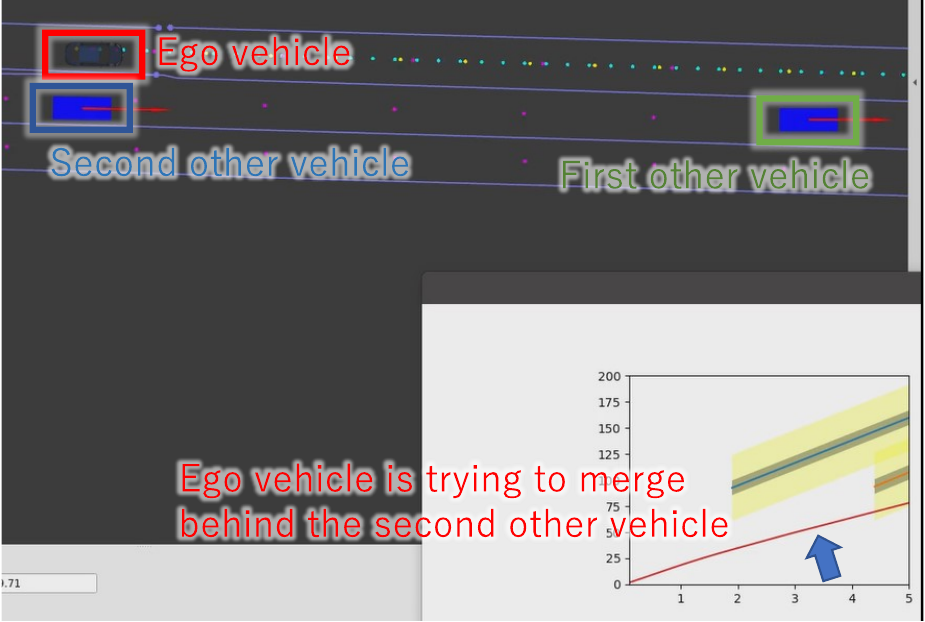}
\caption{Intermediate result of the speed planning during processing when the second vehicle is accelerating.}
\label{fig:real_test_process}
\end{figure}

The test course is same as the aforementioned simulation experiment.
The state of the other vehicles such as speed and acceleration is obtained with VBOX instead of external sensors.
We show the picture of the experiment in progress in Fig. .
Fig.  shows the process of the speed planning.
We confirmed that the ego vehicle behaves the expected merging motion.

\section{Conclusion}
This paper presented a deterministic MDP based speed planning algorithm for merging in highway ramps.
Though the state space is even huge, the calculation can be in real-time by designing the appropriate model and introducing some calculation techniques. We confirmed the basic advantage of our scheme compared with the simple IDM based planner.
Furthermore, we conducted the real vehicle test and the planner could have generated suitable behaviours in response to acceleration and deceleration of other vehicles. 

We need to conduct a detailed evaluation for future work. In addition, we are interested in developing an algorithm that decides the merging position, expansion of the scheme to other scenarios such as urban roads, and the further consideration of the reward function.


\end{document}